\documentclass[journal,twoside,web]{ieeecolor}

\usepackage{tmi}
\usepackage{cite}
\usepackage{amsmath,amssymb,amsfonts}
\usepackage{graphicx}
\usepackage{textcomp}

\usepackage[breaklinks,hidelinks]{hyperref}
\usepackage{url} 

\usepackage{cleveref}

\usepackage{stmaryrd}
\usepackage{booktabs}
\usepackage{tabularx}
\usepackage{dirtytalk}
\usepackage{multirow}
\usepackage{adjustbox}
\usepackage{bm}
\usepackage{xurl}

\usepackage[acronym]{glossaries}
\newacronym{a4c}{A4C}{Apical 4 Chamber}
\newacronym{adm}{ADM}{Ablated Diffusion Model}
\newacronym{ai}{AI}{Artificial Intelligence}
\newacronym{ann}{ANN}{Artificial Neural Network}
\newacronym{bert}{BERT}{Bidirectional Encoder Representations from Transformers}
\newacronym{cdm}{CDM}{Cascaded Diffusion Model}
\newacronym{cnn}{CNN}{Convolutional Neural Network}
\newacronym{ct}{CT}{Computed Tomography}
\newacronym{ddim}{DDIM}{Denoising Diffusion Implicit Model}
\newacronym[longplural={Denoising Diffusion Probabilistic Models},plural={DDPMs}]{ddpm}{DDPM}{Denoising Diffusion Probabilistic Model}
\newacronym{dnn}{DNN}{Deep Neural Network}
\newacronym{dyn}{Dynamic}{EchoNet-Dynamic}
\newacronym{lvh}{LVH}{EchoNet-LVH}
\newacronym{dtgn}{D'ARTAGNAN}{Deep ARtificial Twin-Architecture GeNerAtive Networks}
\newacronym{ed}{ED}{End-Diastolic}
\newacronym{es}{ES}{End-Systolic}
\newacronym{edm}{EDM}{Elucidated Diffusion Model}
\newacronym{ef}{EF}{Left Ventricular Ejection Fraction}
\newacronym{ffn}{FFN}{Feed-Forward Neural Network}
\newacronym{fid}{FID}{Fréchet Inception Distance}
\newacronym{fvd}{FVD}{Fréchet Video Distance}
\newacronym{fps}{fps}{frames per second}
\newacronym[longplural={Generative Adversarial Networks},plural={GANs}]{gan}{GAN}{Generative Adversarial Network}
\newacronym{gru}{GRU}{Gated Recurrent Unit}
\newacronym{is}{IS}{Inception Score}
\newacronym{ldm}{LDM}{Latent Diffusion Model}
\newacronym{lidm}{LIDM}{Latent Image Diffusion Model}
\newacronym{lvef}{LVEF}{Left Ventricle Ejection Fraction}
\newacronym{lstm}{LSTM}{Long Short-Term Memory}
\newacronym{lvdm}{LVDM}{Latent Video Diffusion Model}
\newacronym{lpips}{LPIPS}{Perceptual Similarity}
\newacronym{mae}{MAE}{Mean Absolute Error}
\newacronym{ml}{ML}{Machine Learning}
\newacronym{mlp}{MLP}{Multi-Layer Perceptron}
\newacronym{mse}{MSE}{Mean Squared Error}
\newacronym{ssim}{SSIM}{Structural Similarity Index}
\newacronym{psnr}{PSNR}{Peak Signal-to-Noise Ratio}
\newacronym{flops}{FLOPs}{Floating-point Operations per Second}
\newacronym{r2}{$R^2$}{Coefficient of Determination}
\newacronym{rmse}{RMSE}{Root mean square error}
\newacronym{ode}{ODE}{Ordinary Differential Equation}
\newacronym{pndm}{PNDM}{Pseudo Numerical methods for diffusion models}
\newacronym{ped}{Pediatric}{EchoNet-Pediatric}
\newacronym{psax}{PSAX}{Parasternal Short Axis}
\newacronym{plax}{PLAX}{Parasternal Long Axis}
\newacronym{rnn}{RNN}{Recurrent Neural Network}
\newacronym{sde}{SDE}{Stochastic Differential Equation}
\newacronym{syn}{Syn}{EchoNet-Synthetic}
\newacronym{sv}{SV}{Systolic Volume}
\newacronym{us}{US}{Ultrasound}
\newacronym{vae}{VAE}{Variational Auto-Encoder}
\newacronym{vdm}{VDM}{Video Diffusion Model}
\newacronym{vqvae}{VQ-VAE}{Vector Quantized Variational Autoencoder}
\newacronym{mri}{MRI}{Magnetic Resonance Imaging}
\newacronym{lifm}{LIFM}{Latent Image Flow Matching}
\newacronym{lvfm}{LVFM}{Latent Video Flow Matching}
\newacronym{avae}{A-VAE}{Adversarial Variational Auto-Encoder}
\newacronym{cfg}{CFG}{Classifier-Free Guidance}
\newacronym{reid}{ReId}{Re-Identification}

\newcommand{\ua}{\uparrow}
\newcommand{\da}{\downarrow}

\def\BibTeX{{\rm B\kern-.05em{\sc i\kern-.025em b}\kern-.08em
    T\kern-.1667em\lower.7ex\hbox{E}\kern-.125emX}}
\begin{document}

\title{EchoFlow: A Foundation Model for Cardiac Ultrasound Image and Video Generation}

\author{\mbox{
    Hadrien Reynaud,
    Alberto Gomez, 
    Paul Leeson,
    Qingjie Meng$^\ast$,
    and Bernhard Kainz$^\ast$, \IEEEmembership{Senior Member, IEEE}
    }
    %
    %
    \thanks{Manuscript received x; 
    }
    \thanks{This work was supported by Ultromics Ltd., the UKRI Centre for Doctoral Training in Artificial Intelligence for Healthcare  (EP / S023283/1) and HPC resources provided by the Erlangen National High Performance Computing Center (NHR@FAU) of the Friedrich-Alexander-Universität Erlangen-Nürnberg (FAU) under the NHR project b180dc. NHR and high-tech agenda Bavaria (HTA) funding is partly provided by federal and Bavarian state authorities. NHR@FAU hardware is partially funded by the German Research Foundation (DFG) - 440719683. Support was also received from the ERC project MIA-NORMAL 101083647 and DFG KA 5801/2-1, INST 90/1351-1, and 512819079.}
    \thanks{H. Reynaud is with the UKRI CDT in AI for Healthcare and Department of Computing, Imperial College London, London, UK (e-mail:hadrien.reynaud19@imperial.ac.uk). Q. Meng is with the School of Computer Science, University of Birmingham, Birmingham, UK and Department of Computing, Imperial College London, London, UK. A. Gomez is with Ultromics Ldt. P. Leeson is with Ultromics Ldt. and University of Oxford. B. Kainz is with the Department of Computing, Imperial College London, London, UK and Friedrich–Alexander University Erlangen–Nürnberg, DE.$^\ast$ Equal supervision.}
}

\maketitle

\begin{abstract}
Advances in deep learning have significantly enhanced medical image analysis, yet the availability of large-scale medical datasets remains constrained by patient privacy concerns. We present EchoFlow, a novel framework designed to generate high-quality, privacy-preserving synthetic echocardiogram images and videos. EchoFlow comprises four key components: an adversarial variational autoencoder for defining an efficient latent representation of cardiac ultrasound images, a latent image flow matching model for generating accurate latent echocardiogram images, a latent re-identification model to ensure privacy by filtering images anatomically,  and a latent video flow matching model for animating latent images into realistic echocardiogram videos conditioned on ejection fraction. We rigorously evaluate our synthetic datasets on the clinically relevant task of ejection fraction regression and demonstrate, for the first time, that downstream models trained exclusively on EchoFlow-generated synthetic datasets achieve performance parity with models trained on real datasets. We release our models and synthetic datasets, enabling broader, privacy-compliant research in medical ultrasound imaging at \url{https://huggingface.co/spaces/HReynaud/EchoFlow}.
\end{abstract}



\section{Introduction}
\label{sec:introduction}

\IEEEPARstart{A}{dvancements} in artificial intelligence and deep learning have significantly benefited a variety of fields in healthcare, such as diagnosis, treatment planning, and patient care~\cite{rajpurkar2022ai,yu2018artificial}. Medical datasets are fundamental for these data-driven deep learning approaches, as they provide the critical information necessary for precise predictions and rigorous analyses~\cite{ouyang2020video-based,rueckert2016learning}. However, the inherently sensitive nature of patient data, coupled with stringent privacy regulations and the challenges of obtaining consent, often limits the accessibility and dissemination of medical datasets. 

Echocardiogram analysis exemplifies these challenges. Echocardiogram, or cardiac ultrasound, plays a pivotal role in assessing cardiac structure and function, as well as diagnosing cardiovascular diseases. Nevertheless, acquiring sufficient echocardiographic data is costly, as the acquisition process heavily depends on the sonographer's expertise and it usually takes years to train qualified sonographers.

Recently, generative models such as \glspl{gan}, \glspl{vae} and diffusion models have emerged as powerful tools for producing high-quality images and videos in computer vision~\cite{brock2019large,child2021very,vandenoord2016pixel,saharia2022photorealistic,kumari2022multi-concept,singer2022make-a-video,rakhimov2020latent,he2022latent}.
Several recent works have successfully employed diffusion models to synthesise echocardiogram images~\cite{stojanovski2023echo,nguyen2024training-free} and videos~\cite{zhou2024heartbeat,chen2024ultrasound,li2024echopulse,reynaud2024echonet-synthetic}. They demonstrate good quality through commonly used quantitative metrics, however, few have evaluated the practical utility of synthetic data in clinical applications. 

In this work, we introduce EchoFlow, a novel and systematic framework to synthesise echocardiogram videos. Clinical downstream tasks are utilised as a measurement to explicitly evaluate the practical utility of our synthetic data. Our framework consists of two main components: data generation and downstream evaluation. In the data generation stage, we first train an \gls{avae} for image reconstruction, using its encoder to define a latent representation of real data and its decoder to generate images from this representation. We then develop latent flow matching within the \gls{avae}-defined latent space to train both the \gls{lifm} and \gls{lvfm} models.
To  safeguard patient privacy and mitigate risks of data leakage, a rigorous de-identification protocol, implemented via a \gls{reid} module, is incorporated. Once these four modules are trained, synthetic echocardiogram datasets are generated by using the \gls{lifm} model to produce individual frames, the \gls{reid} module for privacy filtering, and the \gls{lvfm} model to generate echocardiogram videos in the latent space, which are subsequently decoded by the \gls{avae} decoder. In the downstream evaluation stage, \gls{ef} regression is used as a clinical downstream task to validate the clinical utility of our synthetic dataset. Specifically, we compare the performance of \gls{ef} regression models trained on our synthetic data against identical models trained on real datasets.

A preliminary version of this work was presented at the MICCAI 2024 conference~\cite{reynaud2024echonet-synthetic}, where a \gls{ldm} was employed for echocardiogram generation. Although~\cite{reynaud2024echonet-synthetic} yielded high-quality synthesised echocardiograms, a gap persisted between the performance of the synthetic data and that of real data in downstream \gls{ef} regression tasks. In contrast, this work thoroughly validates a broad range of \gls{vae} techniques and generative models, leading to a comprehensive data generation pipeline that produces synthetic echocardiograms with performance closely matching that of real data in clinical applications. Moreover, our work establishes a foundational framework for medical data generation, offering valuable insights into creating synthetic data with genuine clinical utility.

\subsubsection*{Contributions} Our main contributions are: 
\begin{itemize}

\item We introduce a systematic and foundational framework that leverages flow matching techniques for echocardiogram synthesis. We are the first to use latent flow matching for cardiac ultrasound generation. 

\item We develop and thoroughly validate a comprehensive data generation pipeline. By optimizing the architecture, scale, and training protocols for latent space learning as well as image and video synthesis, our proposed pipeline produces high-fidelity echocardiogram videos.

\item We introduce clinically relevant downstream tasks to assess the practical utility of our synthetic datasets. Importantly, our work is the first to generate synthetic echocardiogram videos that can achieve performance parity with real data in these tasks.

\item We perform an extensive experimental analysis across three publicly available datasets. Our evaluations include quantitative and qualitative assessments of synthetic echocardiograms and downstream \gls{ef} regression performance. Additionally, we compare the proposed method with state-of-the-art echocardiogram generation methods and conduct a detailed ablation study of the data generation sampling process.

\end{itemize}

\section{Related Works}

\subsection{Generative Models for Image Synthesis}

Image synthesis in computer vision has gained significant attention in recent years. \glspl{gan}~\cite{goodfellow2014generative} have been widely used for image synthesis, as they enable efficient sampling of high-resolution images with good perceptual quality~\cite{brock2019large}. However, \glspl{gan} often suffer from mode collapse and unstable training~\cite{arjovsky2017wasserstein}. \glspl{vae}~\cite{kingma2014auto-encoding} also support efficient image synthesis~\cite{child2021very} but they usually tend to produce blurry results. 

Recently, diffusion models have emerged as a leading approach for generation tasks~\cite{ho2020denoising,rombach2022high-resolution,saharia2022photorealistic}. They offer stable training and excellent scalability, achieving high sample quality and broad mode coverage~\cite{dhariwal2021diffusion,saharia2022photorealistic}. Despite these strengths, evaluating and optimizing diffusion models in pixel space results in slow inference speeds and high training costs. To address these issues, researchers have moved the denoising process into latent spaces. For example, \cite{rombach2022high-resolution} trains a diffusion model in the latent space of a powerful pre-trained \gls{avae}. This approach paved the way for many subsequent works, including the large-scale implementations known as Stable Diffusion~\cite{compvis-stable-diffusion-v1-4,stabilityai-stable-diffusion-2-1,stabilityai-stable-diffusion-3.5-large}. Many following works have fine-tuned the pre-trained Stable Diffusion weights for specific tasks \cite{brooks2023instructpix2pix,kumari2022multi-concept}. 
Compared to diffusion models~\cite{ho2020denoising,rombach2022high-resolution} 
recent flow matching approaches offer greater data and compute efficiency~\cite{lipman2023flow,lee2024improving,esser2024scaling}. Flow matching redefines the forward process as a direct path between the data distribution and a standard normal distribution. This offers a more straightforward transition from noise to data.

\subsection{Generative Models for Video Synthesis}

For video generation, popular image synthesis models have been adapted, such as \glspl{gan} and \glspl{vae}~\cite{singer2022make-a-video,rakhimov2020latent}. 
Diffusion models have demonstrated reasonable performance on videos with low temporal and spatial resolutions~\cite{ho2022video}. They have also achieved high-definition quality for longer samples when, \emph{e.g.}, conditioned on text inputs~\cite{ho2022imagen,khachatryan2023text2video-zero,singer2022make-a-video}.
Although these methods exhibit outstanding modelling capabilities, they often suffer from excessive computational requirements. To improve tractability, recent work have explored \glspl{ldm}~\cite{rombach2022high-resolution} for video generation~\cite{blattmann2023align,he2022latent}. VideoLDM \cite{blattmann2023align} extends \glspl{ldm} to high-resolution video generation by converting pre-trained image \glspl{ldm} into video generators through the insertion of temporal layers. \gls{lvdm} \cite{he2022latent} proposes a hierarchical diffusion model that operates in the video latent space, enabling the generation of long video sequences through a secondary model. More recently, a few studies have begun exploring flow matching techniques for more efficient video generation~\cite{jin2025pyramidal}.

\subsection{Ultrasound Generation}

In the field of ultrasound generation, research has followed two primary directions. Some works focus on physics-based simulators \cite{jensen2004simulation,shams2008real-time}. Others focus on generating individual ultrasound images with generative models. GAN-based methods have been well studied for ultrasound image synthesis. These approaches typically condition their models on complementary imaging modalities such as MRI, CT \cite{teng2020interactive,tomar2021content-preserving}, or simulated images \cite{gilbert2021generating,tiago2022data}. More recent works take advantage of diffusion models to achieve stable and controllable ultrasound image generation~\cite{stojanovski2023echo,nguyen2024training-free}. Lately, ultrasound video generation is becoming a popular research area~\cite{zhou2024heartbeat,chen2024ultrasound,li2024echopulse}. \cite{liang2022sketch} presents a motion-transfer-based method for pelvic ultrasound videos, while \cite{reynaud2022d’artagnan} introduces a causal model for echocardiogram video generation. \cite{reynaud2023feature-conditioned} further develops a diffusion model-based method for synthesizing echocardiogram videos. Although this approach produces high-quality ultrasound videos, it requires extensive sampling times. More recently, \cite{reynaud2024echonet-synthetic} leveraged latent diffusion models for faster and more temporally consistent echocardiogram generation.

\begin{figure*}[!t]
    \centering
    \includegraphics[width=1.0\linewidth]{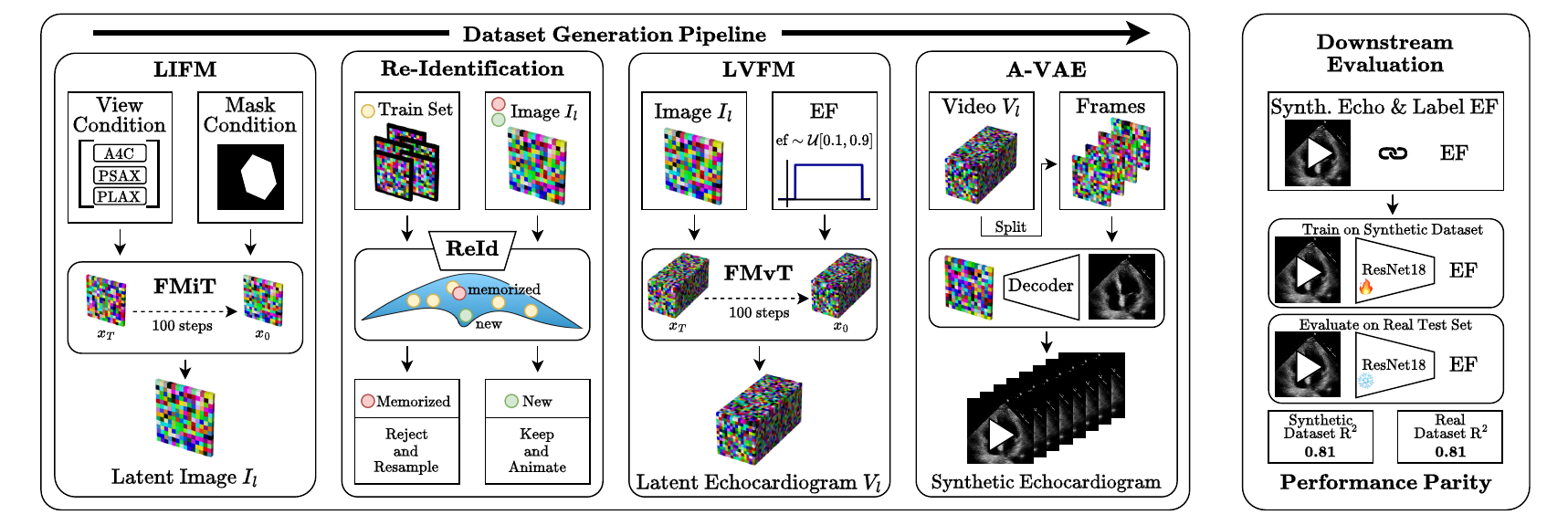}
    \caption{Our EchoFlow framework. From left to right: The image generation model (LIFM), the privacy filter (Re-Identification), the video generation model (LVFM), the decoding stage (A-VAE) and our downstream evaluation. For each step, inputs are shown at the top, process in the middle and output at the bottom.}
    \label{fig:framework}
\end{figure*}
\subsection{Privacy-preserving Generation}

Generative model such as diffusion-based and flow-based models, have a tendency to memorize their training set~\cite{carlini2023extracting,chen2024generative}, which raises privacy concerns.
Two approaches have emerged to try and solve this problem, differential privacy~\cite{dombrowski2023quantifying,cao2021don’t} and \gls{reid}~\cite{dar2024unconditional}. For differential privacy, noise is added to the real samples during training, to prevent memorization. In \gls{reid}, generated samples are filtered during inference to detect any strong match between generated and real samples.
Here, we focus on \gls{reid}, as it maintains model performance.

\section{Methods}

The goal of this work is to achieve performance parity on downstream tasks between a model trained exclusively on synthetic data, and the same model trained on real data. In this section, we present the overarching framework we use to achieve this goal, as well as each individual component.

\subsection{Framework}
Our core components are (1) an \gls{avae}, used to define a domain-specific latent space, (2) a latent image generation model that can generate latent representation of cardiac ultrasound images, (3) a latent \gls{reid} model that can detect when a generated image is anatomically similar to an image from the training set and (4) a latent video generation model that can animate a latent cardiac ultrasound image into a latent video. The generated latent images and videos are decoded back to pixel space using the decoder of the \gls{avae}.
By combining these four blocks, we enable the generation of synthetic datasets that retain the data quality of the real dataset on which the models were trained on, while preventing any real patient data leakage. 
We present our inference and evaluation pipeline in \Cref{fig:framework}.

At training time, it is essential to start by training the \gls{avae} (\Cref{sec:avae}), as it defines the latent space in which all other models will operate. The three other models are independent.

At inference time, we start by sampling a latent echocardiogram image, that acts as a candidate anatomy. This candidate anatomy is sent to our \gls{reid} model, which compares that anatomy with all existing training anatomies. If the similarity score is too high, the latent image is rejected and a new one is generated. Otherwise, the latent anatomy is used to condition the latent video generation model, which animates the heart with a given \gls{ef} score. This produces a sequence of frames, which are then decoded back into pixel space by using the \gls{avae} decoder.

\subsection{Adversarial Variational Auto-Encoder}
\label{sec:avae}

As training and sampling generative models at native resolution can be prohibitively expensive, we first compress the data into latent representation using \glspl{avae}~\cite{rombach2022high-resolution}.
This projects the input image to a 2D latent space. The 2D space allows the model to retain positional information effortlessly by maintaining the relative position of the image features.

Additionally to remedy blurry reconstruction problem, the decoder is shifted from a decoding task via \gls{mse} loss to a generation task with a strong prior, by introducing an adversarial objective in the loss formulation of the decoder. This loss forces the decoder to fill-in missing high frequency details that are lost during the encoding process, and thus not present in the latent space.

However, relying on a 2D latent space prevents the \gls{avae} from generating new samples from random noise, because the decoder network requires structured latent information, which cannot be obtained from simple numerical distributions. Additionally, the introduction of the adversarial loss makes the training less stable, as the decoder behaves like a \gls{gan}~\cite{arjovsky2017wasserstein}.

Existing general-purpose \glspl{avae}~\cite{rombach2022high-resolution,blackforest-flux1,stabilityai-stable-diffusion-3.5-large} can efficiently embed diverse images into a latent space, potentially removing the need for training a custom model. However, because they are trained on general image corpora, they lack the specialisation needed to fully capture echocardiography data, as shown in \Cref{table:avae}. 
Therefore, we train our own \glspl{avae} from scratch, following best practices~\cite{rombach2022high-resolution,blackforest-flux1}.

The \gls{avae} forward pass is as follows. 
An encoder network $E_{\psi}(\cdot)$ maps an input image $x_i$ to a latent space distribution, by predicting both the mean $\mu_i$ and the variance $\sigma_i$ for that image.
The latent variable $z_i$ is sampled using the reparameterization trick (\cref{eq:vae_reparam}). $z_i$ is then be decoded back to the image space using a decoder network $D_{\phi}(\cdot)$.
Formally, 
\begin{align}
    \mu_i, \sigma_i &= E_{\psi}(x_i), \label{eq:avae_encode} \\ 
    z_i &= \mu_i + \sigma_i \odot \epsilon_i, \quad \epsilon_i \sim \mathcal{N}(0,x_i), \label{eq:vae_reparam} \\ 
    \hat{x}_i &= D_{\phi}(z_i). \label{eq:avae_decode}
\end{align}

We train $E_{\psi}(\cdot)$ and $D_{\phi}(\cdot)$ end-to-end through a reconstruction loss, an adversarial loss and a KL-divergence regularization term.
The individual loss terms are given by:
\begin{align}
    \mathcal{L}_{\text{rec}} &= \frac{1}{N} \sum_{i=1}^N \left\| x_i - \hat{x}_i \right\|^2, \\
    \mathcal{L}_{\text{adv}} &= -\mathbb{E}_{z \sim q_\theta(z|x)}\left[\log A\bigl(D_\phi(z)\bigr)\right], \\
    \mathcal{L}_{\text{KL}} &= \frac{1}{2} \sum_{j=1}^J \left(1 + \log(\sigma_j^2) - \mu_j^2 - \sigma_j^2\right), \\
    \mathcal{L}_{\text{VAE}} &= \mathcal{L}_{\text{rec}} + \lambda\cdot\mathcal{L}_{\text{adv}} + \gamma\cdot\mathcal{L}_{\text{KL}}.\label{eq:vae_final_loss}
\end{align}

Here, $\mathcal{L}_{\text{rec}}$ is the reconstruction loss computed using an \gls{mse} between the input image $x_i$ and the reconstructed image $\hat{x}_i$. $N$ is the number of samples in the batch.
$\mathcal{L}_{\text{adv}}$ is the adversarial loss, where the discriminator $A(\cdot)$ is a neural network trained end-to-end with the \gls{avae}, which assesses the quality of the generated image $D_{\phi}(z)$, and $q_\theta(z|x)$ is the encoder's output distribution.
$\mathcal{L}_{\text{KL}}$ is the KL-divergence regularization term over the latent space $z$, and $J$ is the latent space dimension.
$\lambda$ and $\gamma$ are hyperparameters used to adjust the weight of each term over the final loss.

\textbf{Normalisation.}
While the KL regularization term forces the latent space distribution to resemble a Gaussian distribution, the resulting latent space is often skewed from such distribution due the precedence of reconstruction quality, expressed by a low value of $\gamma$ in \Cref{eq:vae_final_loss}. Nonetheless, having a Gaussian distributed latent space is desirable for generative models, as they are easier to train over Gaussian distributed data.
As such, during inference, we normalise any sampled latent $z_i$ channel-wise, based on statistics computed over the corresponding latent training set.
Formally,
\begin{align}
    \mu_k &= \frac{1}{N} \sum_{i=1}^{N} [z_i]_k, \quad \text{for } k=1,\dots,K, \\
    \sigma_k &= \sqrt{\frac{1}{N} \sum_{i=1}^{N} \left([z_i]_k - \mu_k\right)^2}, \quad \text{for } k=1,\dots,K, \\
    [\hat{z}_i]_k &= \frac{[z_i]_k - \mu_k}{\sigma_k}, \quad \text{for } k=1,\dots,K\label{eq:vae_norm}
\end{align}

where $[z_i]_k$ denotes the $k^{th}$ channel of the latent code $z_i$, and the operations in \Cref{eq:vae_norm} are applied channel-wise.

\subsection{Latent Flow Matching Models}
\label{sec:flow_matching}

Flow matching is a deterministic  framework that learns a continuous mapping from a simple distribution to the target data by approximating a velocity field. Rather than relying on a stochastic reverse process like diffusion models, flow matching trains a neural network to follow a well defined interpolation path between noise and data.

%
Let $p_{data}$ denote the data distribution and $p_{prior}$ a simple distribution (\emph{e.g.} Gaussian). We define a family of intermediate distributions $\{p_t(\mathbf{x})\}_{t\in[0,1]}$ such that $p_0 = p_{data}$ and $p_1 = p_{prior}$. We use linear interpolation
\begin{equation}
    \mathbf{x}_t = (1-t)\,\mathbf{x}_0 + t\,\mathbf{x}_1,
\end{equation}
with $\mathbf{x}_0\sim p_{data}$ and $\mathbf{x}_1\sim p_{prior}$. The corresponding ground-truth velocity along the path is given by 
\begin{equation}
    u(\mathbf{x}_0,\mathbf{x}_1) = \frac{d\mathbf{x}_t}{dt} = \mathbf{x}_1 - \mathbf{x}_0,
\end{equation}
where $u(\mathbf{x}_0,\mathbf{x}_1)$ represents the instantaneous velocity vector that drives the interpolation from the data point $\mathbf{x}_0$ to the noise sample $\mathbf{x}_1$.

Training involves learning a time-dependent vector field $v_{\theta}(\mathbf{x},t)$ by minimizing the regression loss:
\begin{equation}
    \mathcal{L}_{\text{FM}} = \mathbb{E}_{t,\mathbf{x}_0,\mathbf{x}_1}\left[\left\|v_{\theta}(\mathbf{x}_t,t) - (\mathbf{x}_1 - \mathbf{x}_0)\right\|^2\right].
\end{equation}

Once trained, new data samples are generated by drawing a sample $\mathbf{x}_1 \sim p_{prior}$. We then discretize the time interval $[0,1]$ into $N$ equal steps with $\Delta t = 1/N$ and apply the Euler ODE sampler to integrate the ODE
\begin{equation}
    \frac{d\mathbf{x}}{dt} = v_{\theta}(\mathbf{x},t),
\end{equation}
backwards from $t=1$ to $t=0$. At each discrete time step, the sample is updated as
\begin{equation}
    \mathbf{x}_{t-\Delta t} = \mathbf{x}_t - \Delta t\, \cdot v_{\theta}(\mathbf{x}_t,t).
\end{equation}
After $N$ steps, the final state $\mathbf{x}_0$ is obtained, which approximates the target data distribution.

Flow Matching is compatible with any backbone architecture.
In this work, we focus on  UNet~\cite{ronneberger2015u-net,ho2020denoising} and Transformers~\cite{vaswani2017attention,peebles2023scalable}. We propose  a Spatial-only model, and a Spatio-Temporal model. 
Our UNet architecture relies on Attention-based ResNet blocks~\cite{schlemper2019attention}. For the Spatio-Temporal UNet, we interleave Spatial and Temporal ResNet blocks and Spatial and Temporal  Attention blocks.

Our Transformer-based architecture follows the implementation from~\cite{peebles2023scalable}. We transform the input image or video into a sequence of patches and add positional (and temporal) embeddings to each patch. The resulting patches are then sent through a sequence of transformer blocks. The output is projected and reshaped back into an image (or video) with the same dimension as the input. For the video, we use Spatio-Temporal transformer blocks which alternate Spatial and Temporal Attention.

\textbf{Conditioning.}
We define two sets of conditionings, one for the image models and one for the video models.
The \gls{lifm} models are conditioned on a segmentation mask and a class index corresponding to the desired view (\gls{a4c}, \gls{psax} or \gls{plax}). This allows us to control the view but also the area covered by the left ventricle. 
The \gls{lvfm} models are conditioned on an anatomy (\emph{i.e.}, an echocardiogram frame) and an \gls{ef} score, shifting the task from pure generation to frame-animation. We ensure during training that none of the generated frames are identical to the conditioning frame, to force the model to disentangle anatomy understanding from the generation task.

To improve the adherence of the generated samples to our conditionings, we rely on \gls{cfg}~\cite{ho2022classifier-free}. The models are trained with a given probability to drop some or all their conditionings, which allows us to sample both a conditional sample and an unconditional sample during inference, and apply \gls{cfg} such that
\begin{equation}
        v_{unc} = v_{\theta}(\mathbf{x}_t,t), \quad
        v_{cond} = v_{\theta}(\mathbf{x}_t,t,c_1,c_2), \\
        \label{eq:cfg_def}
\end{equation}
\begin{equation}
        v_{\text{cfg}} = v_{unc} - \lambda_{\text{cfg}} \cdot ( v_{cond} - v_{unc} ) \\
\end{equation}
\begin{equation}
        \mathbf{x}_{t-\Delta t} = \mathbf{x}_t - \Delta t\, \cdot v_{\text{cfg}},
\end{equation}
where $c_1,c_2$ are conditionings (\emph{e.g.}, an \gls{ef} score and an anatomy), and $\lambda_{\text{cfg}}$ is the \gls{cfg} scale. If $\lambda_{\text{cfg}}=0$, the sampling is unconditioned, while setting $\lambda_{\text{cfg}}=1$ ignores \gls{cfg}.

\subsection{Re-Identification}
\label{subsec:reid}
\Gls{reid} is applied as a filter on generated samples, and does not directly interact with any of the generative models. This is extremely powerful as it does not compromise the generative model performance, while still enforcing privacy. Nonetheless, it also means that only the filtered data is privacy compliant, not the generative models themselves. For the objective of this work, this is a suitable solution.

Our \gls{reid} models operate over the latent space, which avoids the expensive operation of projecting latent images back to pixel space, and makes training and inference computationally efficient. 
%
We train \Gls{reid} with a contrastive learning approach. The objective is to learn an encoder $\text{Enc}(\cdot)$ that projects an input image $I$ into a 1D latent space $l$, capturing its anatomy.

At each training step, two videos $V_a$ and $V_b$ are selected. From $V_a$, two random frames $I_{a_1}$ and $I_{a_2}$ are sampled, while from $V_b$ a single frame $I_b$ is sampled. The frames are then paired to define $\text{pair}_{\text{pos}} = \{ I_{a_1},\, I_{a_2} \}$ and $\text{pair}_{\text{neg}} = \{ I_{a_1},\, I_{b} \}$.
The encoder projects each image into the latent space $l=\text{Enc}(I)$.
The training objective is to ensure that the latent representations of the positive pair, $l_{a_1}=\text{Enc}(I_{a_1})$ and $l_{a_2}= \text{Enc}(I_{a_2})$, are close, while the representation of the negative pair, $l_{a_1}$ and $l_{b} = \text{Enc}(I_b)$, are pushed apart. This is achieved by optimizing a contrastive loss function.
Formally, we define the sigmoid function $\mathcal{S}(x) = \frac{1}{1+e^{-x}}$, and the training procedure
\begin{gather}
\hat{l}_{a_1} = \mathcal{S}(l_{a_1}), \quad \hat{l}_{a_2} = \mathcal{S}(l_{a_2}), \quad \hat{l}_b = \mathcal{S}(l_b),\\
\Delta_{\text{pos}} = \left|\hat{z}_{a_1} - \hat{l}_{a_2}\right|, \quad \Delta_{\text{neg}} = \left|\hat{l}_{a_1} - \hat{l}_b\right|,\\
\hat{y}_{\text{pos}} = W\,\Delta_{\text{pos}} + b, \quad \hat{y}_{\text{neg}} = W\,\Delta_{\text{neg}} + b,\\
\mathcal{L}_{\text{reid}} = \text{BCE}\big(\hat{y}_{\text{pos}}, 1\big) + \text{BCE}\big(\hat{y}_{\text{neg}}, 0\big).
\end{gather}
Here, $W$ and $b$ are learnable parameters trained end-to-end with $\text{Enc}(\cdot)$ and `BCE' is the Binary Cross-Entropy loss. 

During inference, the trained \gls{reid} encoder $\text{Enc}(\cdot)$ is applied to an input image $I$ to obtain an anatomical representation.
Since the training set comprises videos and all frames within a video have very similar anatomical embeddings, only the first encoded frame $l_v^{1} = \text{Enc}(I_v^{1})$ from each video $v$ is used to represent a whole video.

To determine the similarity between the input image and the training set, the Pearson correlation coefficient is computed between the latent representation $l$ of the input and all training representation $l_v^{1}$:
\begin{equation}
\rho\big(l, l_v^{1}\big) = \frac{\operatorname{cov}\big(l, l_v^{1}\big)}{\sqrt{\operatorname{var}(l)\,\operatorname{var}\big(l_v^{1}\big)}}.
\end{equation}
The maximum correlation value across all training videos is then selected:
\begin{equation}
\rho_{\text{max}} = \max_{v}\; \rho\big(l, l_v^{1}\big).
\end{equation}
Finally, this maximum similarity score $\rho_{\text{max}}$ is compared against a precomputed privacy threshold $\tau$. If $\rho_{\text{max}} \geq \tau$, the input image is considered too similar to the training data, and flagged as not respecting privacy guarantees.
The privacy threshold $\tau$ is determined by comparing the similarities between the real training and real validation sets. Let $\{l_i^{1,\text{val}}\}$ denote the embeddings from the validation set first frames. The Pearson correlation coefficient is computed between each training embedding and every validation embedding, forming a correlation matrix $C$. For each training sample $v$, the maximum correlation value is extracted:
\begin{equation}
\rho_v^{\text{max}} = \max_{i} \, \rho\big(l_v^{1}, l_i^{1,\text{val}}\big).
\end{equation}
The threshold $\tau$ is then set as the $p^{\text{th}}$ percentile of these maximum correlations:
\begin{equation}
\tau = \text{percentile}\Bigl(\{\rho_v^{\text{max}} : v \in \text{training set}\},\, p\Bigr),
\end{equation}
where $p$ is a predefined cutoff percentage set to $95\%$. This approach ensures that $\tau$ reflects the inherent similarity between training and validation samples, providing a reliable benchmark for assessing anatomical similarity.

\subsection{Ejection Fraction Regression}
We use downstream task performance as the main metric for this work, as we believe it is the best way to demonstrate the quality of a synthetic dataset. In the context of echocardiograms, \gls{ef} regression is the most common task~\cite{ouyang2020video-based,reynaud2021ultrasound}. The goal of \gls{ef} regression is to estimate a continuous scalar value from a sequence of input frames.
The EF regression task is performed on datasets of the form $\mathcal{D}_{ef} = \{ (V, y) \}$ that consist of a list of pairs of video \(V\) and their corresponding ground-truth \gls{ef} score \(y\). During training, a video \(V\) is fed into a regression model \(\text{Reg}(\cdot)\) to obtain an estimate $\hat{y} = \text{Reg}(V)$.
The model is optimised by minimizing the \gls{mse} loss between the true \gls{ef} score \(y\) and the estimated \gls{ef} score \(\hat{y}\) such that $\mathcal{L}_{\text{reg}} = \left(y - \hat{y}\right)^2$.
This loss encourages the regression model to accurately capture the cardiac features present in the video, leading to reliable predictions of the \gls{ef}.

\section{Experiments}
\label{sec:experiments}

\subsection{Experimental setups}

\subsubsection{Data}

To facilitate reproducibility, we rely exclusively on public echocardiogram datasets  \gls{dyn}~\cite{ouyang2020video-based}, \gls{ped}~\cite{reddy2023video-based} and \gls{lvh}~\cite{stanford-echonet-lvh}, as real datasets. 
\gls{dyn} contains exclusively \gls{a4c} views, and we adopt its original training, validation, and testing split.
\gls{ped} contains both \gls{a4c} and \gls{psax} views, so we split it into two separate datasets \gls{ped}~(\gls{a4c}) and \gls{ped}~(\gls{psax}). Both \gls{ped} datasets are originally split into 10 folds, and we arbitrarily use the first 8 folds for training, the $9^{\text{th}}$ for validation and the $10^{\text{th}}$ for testing.
\gls{lvh} contains exclusively \gls{plax} views, and we use its original splits.
Every echocardiogram in the \gls{dyn} and \gls{ped} datasets comes with an expert-labelled \gls{ef} score and two segmented frames, corresponding to an \gls{es} and \gls{ed} key frame.
The \gls{lvh} dataset comes with tracings of the left ventricle. We use these tracings to estimate the \gls{es} and \gls{ed} volumes (ESV and EDV) through the Teichholz formula~\cite{teichholz1976problems}. This allows us to calculate an approximate \gls{ef} score as $\text{EF}=\frac{\text{EDV} - \text{ESV}}{\text{EDV}}\times100$. 
Our image generation models take a segmentation mask as a condition. To allow each frame of \gls{dyn} and \gls{ped} to have a paired segmentation, we train a DeepLabV3 segmentation model~\cite{chen2017rethinking}, and use it to segment every unlabelled frame. We use empty masks for \gls{lvh} as it does not have segmentation labels.
The size of each dataset and each split is reported in \Cref{table:echoflow}.
We also rely on TMED2~\cite{huang2022tmed} for \gls{avae} experiments. For TMED2, we disregard any split and labels and build a large database of ultrasound images by merging all available images.

\subsubsection{Evaluation metrics}
To evaluate the performance of \glspl{avae}, we compute both reconstruction metrics and generative metrics. Reconstruction metrics are evaluated pair-wise between an original image and its reconstructions. Generative metrics are computed dataset-wise between the real datasets and the reconstructed datasets.
For the \gls{lifm} and the \gls{lvfm}, we assess the image generation quality using \gls{fid} and \gls{is} and video generation quality using \gls{fvd}. We also measure model efficiency in terms of number of parameters, \gls{flops} and inference time. Beyond these metrics, we use a clinical downstream task as a measurement to ensures that synthetic datasets can effectively support medical imaging research and clinical translation. Specifically, we train individual \gls{ef} regression models on every real dataset and its exclusively synthetic counterpart, and systematically assess performance on the real test datasets. We use the \gls{r2}, \gls{mae} and \gls{rmse} metrics to assess the precision of the \gls{ef} regression models.

\subsection{Adversarial Variational Auto-Encoders}
\subsubsection{Data preparation}

The \glspl{avae} training needs a large collection of images. To build it, we sample 1 in 5 frames from every echocardiogram video in our datasets' training split. Then, to increase robustness and generalization capabilities, we use the TMED2~\cite{huang2022tmed} echocardiogram collection on top of our existing video datasets. This combination leads to a total of $1\,312\,623$ training frames, all with resolution $112\times112\times3$.

\subsubsection{Hyperparameters}

All our \glspl{avae} rely on an encoder-decoder architecture, based on a succession of convolutional ResNet and downsampling blocks.
We train three variants, where we vary the depth of the encoder and decoder which impacts the spatial compression factor, as well as the number of latent space channels. We define 
(1) \gls{avae}-4f8, with $4$ latent channels and a spatial compression factor of $8$, resulting in a latent space of $14\times14\times4$, 
(2) \gls{avae}-4f4, with $4$ channels, compression $4\times$, and a $28\times28\times4$ latent space, and 
(3) \gls{avae}-16f8, with $16$ channels, compression $8\times$, and a  $14\times14\times16$ latent space. Our models cover three latent space resolutions, which is the main factor we want to evaluate.

\subsubsection{Training}

Our \glspl{avae} are trained for 10.5 days on $8\times$A40 GPUs, resulting in 2000 GPU-hours per \gls{avae}, and 300 epochs for each. The batch size is fixed to $32$ per GPU, with a gradient accumulation of $2$, resulting in a total batch size of $512$. The learning rate is set to $1e-3$ and the adversarial loss starts after 30 epochs of training, in order to let the model learn from the reconstruction loss first.

\subsubsection{Baselines}

We compare the performance of our \glspl{avae} with four general \glspl{avae}~\cite{rombach2022high-resolution,stabilityai-stable-diffusion-2-1,stabilityai-stable-diffusion-3.5-large,blackforest-flux1} and one echocardiogram-specific \gls{avae}~\cite{reynaud2024echonet-synthetic} and report the results in \Cref{table:avae}.
All the general \glspl{avae} are trained on general image dataset. SD 1.4 \gls{avae}~\cite{rombach2022high-resolution} is the first widely used \gls{avae}, and SD 2.1 \gls{avae}~\cite{stabilityai-stable-diffusion-2-1} and SD 3.5 \gls{avae}~\cite{stabilityai-stable-diffusion-3.5-large} are its successors, which rely on very similar architectures and data mix. FLUX.1 \gls{avae}~\cite{blackforest-flux1} is a recent state-of-the-art general \gls{avae} that excels at image reconstruction. 
EchoSyn \gls{avae}~\cite{reynaud2024echonet-synthetic} is  trained exclusively on an echocardiogram-specific data mix.

\subsubsection{Evaluation}

We compare the performance all the \glspl{avae} by reconstructing the entire real datasets and evaluating both reconstructions and generative metrics. Results are presented in \Cref{table:avae}, where we show all the metrics for \gls{dyn} and only our selected models metrics for \gls{ped} and \gls{lvh}.

\begin{table*}[t!]
\centering
\caption{Performance of the \gls{vae}. Best values are in bold, second best are underlined.}
\label{table:avae}
\resizebox{2.0\columnwidth}{!}{
\begin{tabular}{lcccccccccccc}
\toprule
&&& \multicolumn{5}{c}{\textbf{Reconstruction Metrics}} && \multicolumn{3}{c}{\textbf{Generative Metrics}} \\
\cmidrule{2-2} \cmidrule{4-8} \cmidrule{10-12}
\textbf{Dynamic (A4C)}& Latent Res. && {MSE}$\downarrow$ & {MAE}$\downarrow$ & {SSIM}$\uparrow$ & {PSNR} (dB)$\uparrow$ & {LPIPS}$\downarrow$ && {FID}$\downarrow$ & {FVD$_{16}$}$\downarrow$ & {IS}$\uparrow$ \\
\cmidrule{2-2} \cmidrule{4-8} \cmidrule{10-12}
SD 1.4 \gls{avae}~\cite{rombach2022high-resolution} 
  & $14\times14\times4$ & & $0.0051_{\pm0.0011}$ 
  & $0.0366_{\pm0.0056}$ 
  & $0.7290_{\pm0.0457}$ 
  & $22.99_{\pm1.02}$ 
  & $0.1169_{\pm0.0113}$ &  & $39.31$ 
  & $141.76$ 
  & $2.52_{\pm0.10}$ \\
SD 2.1 \gls{avae}~\cite{stabilityai-stable-diffusion-2-1}  
  & $14\times14\times4$ & & $0.0043_{\pm0.0009}$ 
  & $0.0340_{\pm0.0050}$ 
  & $0.7597_{\pm0.0418}$ 
  & $23.72_{\pm0.95}$ 
  & $0.1021_{\pm0.0099}$ &  & $30.04$ 
  & $174.22$ 
  & $\underline{2.54}_{\pm0.09}$ \\
SD 3.5 \gls{avae}~\cite{stabilityai-stable-diffusion-3.5-large}
  & $14\times14\times16$ & & $0.0028_{\pm0.0006}$ 
  & $0.0280_{\pm0.0046}$ 
  & $0.8272_{\pm0.0340}$ 
  & $25.62_{\pm1.09}$ 
  & $0.0900_{\pm0.0075}$ &  & $33.61$ 
  & $\underline{89.41}$ 
  & $2.53_{\pm0.10}$ \\
FLUX.1 \gls{avae}~\cite{blackforest-flux1}
  & $14\times14\times16$ & & ${0.0017}_{\pm0.0004}$ 
  & $0.0216_{\pm0.0038}$ 
  & $0.8928_{\pm0.0276}$ 
  & $27.87_{\pm1.21}$ 
  & $0.0409_{\pm0.0060}$ &  & $5.19$ 
  & $95.78$ 
  & $\textbf{2.56}_{\pm0.07}$ \\
EchoSyn \gls{avae}~\cite{reynaud2024echonet-synthetic} 
  & $14\times14\times4$ & & $0.0034_{\pm0.0008}$ 
  & $0.0305_{\pm0.0050}$ 
  & $0.7999_{\pm0.0444}$ 
  & $24.79_{\pm1.08}$ 
  & $0.0756_{\pm0.0101}$ &  & $12.55$ 
  & $90.16$ 
  & $2.37_{\pm0.07}$ \\
\gls{avae} 4f8 (ours)
  & $14\times14\times4$ & & $0.0034_{\pm0.0008}$ 
  & $0.0301_{\pm0.0050}$ 
  & $0.7996_{\pm0.0442}$ 
  & $24.87_{\pm1.14}$ 
  & $0.0787_{\pm0.0101}$ &  & $13.76$ 
  & $119.22$ 
  & $2.37_{\pm0.08}$ \\
\gls{avae} 4f4 (ours)
  & $28\times28\times4$ & & $\textbf{0.0010}_{\pm0.0003}$ 
  & $\textbf{0.0170}_{\pm0.0034}$ 
  & $\textbf{0.9244}_{\pm0.0240}$ 
  & $\textbf{30.26}_{\pm1.42}$ 
  & $\underline{0.0346}_{\pm0.0065}$ &  & $\textbf{3.62}$ 
  & $\textbf{61.86}$ 
  & $2.48_{\pm0.07}$ \\
\gls{avae} 16f8 (ours)
  & $14\times14\times16$ & & $\underline{0.0010}_{\pm0.0003}$ 
  & $\underline{0.0175}_{\pm0.0036}$ 
  & $\underline{0.9173}_{\pm0.0264}$ 
  & $\underline{30.09}_{\pm1.45}$ 
  & $\textbf{0.0345}_{\pm0.0070}$ &  & $\underline{4.10}$ 
  & $104.94$ 
  & $2.46_{\pm0.07}$ \\
\midrule
\textbf{Pediatric (A4C)}& Latent Res. && {MSE}$\downarrow$ & {MAE}$\downarrow$ & {SSIM}$\uparrow$ & {PSNR} (dB)$\uparrow$ & {LPIPS}$\downarrow$ && {FID}$\downarrow$ & {FVD$_{16}$}$\downarrow$ & {IS}$\uparrow$ \\
\cmidrule{2-2} \cmidrule{4-8} \cmidrule{10-12}
\gls{avae} 4f4 (ours)
  & $28\times28\times4$ & & $\textbf{0.0003}_{\pm0.0002}$ 
  & $\textbf{0.0090}_{\pm0.0030}$ 
  & $\textbf{0.9690}_{\pm0.0133}$ 
  & $\textbf{35.79}_{\pm2.23}$ 
  & $\textbf{0.0266}_{\pm0.0086}$ &  & $1.81$ 
  & $\underline{6.00}$ 
  & $2.93_{\pm0.00}$ \\
\gls{avae} 16f8 (ours)
  & $14\times14\times16$ & & $\underline{0.0003}_{\pm0.0002}$ 
  & $\underline{0.0097}_{\pm0.0035}$ 
  & $\underline{0.9570}_{\pm0.0217}$ 
  & $\underline{35.59}_{\pm2.44}$ 
  & $\underline{0.0293}_{\pm0.0115}$ &  & $\underline{1.69}$ 
  & $7.94$ 
  & $2.96_{\pm0.00}$ \\
\midrule
\textbf{Pediatric (PSAX)}& Latent Res. && {MSE}$\downarrow$ & {MAE}$\downarrow$ & {SSIM}$\uparrow$ & {PSNR} (dB)$\uparrow$ & {LPIPS}$\downarrow$ && {FID}$\downarrow$ & {FVD$_{16}$}$\downarrow$ & {IS}$\uparrow$ \\
\cmidrule{2-2} \cmidrule{4-8} \cmidrule{10-12}
\gls{avae} 4f4 (ours)
  & $28\times28\times4$ & & $\textbf{0.0003}_{\pm0.0002}$ 
  & $\textbf{0.0095}_{\pm0.0033}$ 
  & $\textbf{0.9680}_{\pm0.0146}$ 
  & $\textbf{35.50}_{\pm2.38}$ 
  & $\textbf{0.0274}_{\pm0.0094}$ &  & $1.86$ 
  & $\underline{5.39}$ 
  & $3.24_{\pm0.12}$ \\
\gls{avae} 16f8 (ours)
  & $14\times14\times16$ & & $\underline{0.0004}_{\pm0.0002}$ 
  & $\underline{0.0102}_{\pm0.0039}$ 
  & $\underline{0.9570}_{\pm0.0225}$ 
  & $\underline{35.27}_{\pm2.60}$ 
  & $\underline{0.0297}_{\pm0.0122}$ &  & $\textbf{1.51}$ 
  & $7.37$ 
  & $3.20_{\pm0.12}$ \\
\midrule
\textbf{LVH (PLAX)}& Latent Res. && {MSE}$\downarrow$ & {MAE}$\downarrow$ & {SSIM}$\uparrow$ & {PSNR} (dB)$\uparrow$ & {LPIPS}$\downarrow$ && {FID}$\downarrow$ & {FVD$_{16}$}$\downarrow$ & {IS}$\uparrow$ \\
\cmidrule{2-2} \cmidrule{4-8} \cmidrule{10-12}
\gls{avae} 4f4 (ours)
  & $28\times28\times4$ & & $\textbf{0.0010}_{\pm0.0003}$ 
  & $\textbf{0.0170}_{\pm0.0032}$ 
  & $\textbf{0.9308}_{\pm0.0203}$ 
  & $\textbf{30.31}_{\pm1.29}$ 
  & $\underline{0.0352}_{\pm0.0059}$ &  & $\underline{4.58}$ 
  & $\textbf{26.22}$ 
  & $2.62_{\pm0.11}$ \\
\gls{avae} 16f8 (ours)
  & $14\times14\times16$ & & $\underline{0.0010}_{\pm0.0003}$ 
  & $\underline{0.0175}_{\pm0.0033}$ 
  & $\underline{0.9234}_{\pm0.0228}$ 
  & $\underline{30.15}_{\pm1.33}$ 
  & $\textbf{0.0351}_{\pm0.0064}$ &  & $\textbf{4.09}$ 
  & $30.08$ 
  & $2.56_{\pm0.10}$ \\
\bottomrule
\end{tabular}
}
\end{table*}

We observe that the FLUX.1 \gls{avae} is the best performing off-the-shelf model, mostly due to its larger latent space dimension that relies on 16 channels instead of 4. Our own models follow a similar trend, where the \gls{avae}-16f8 model performs substantially better than the \gls{avae}-4f8 model. With the \gls{avae}-4f4 model, we maintain the same total compression factor as with the \gls{avae}-16f8, but trade the number of channels for a higher spatial resolution. This results in our best performing model, where more of the spatial information is retained.
Our \gls{avae}-4f4 and \gls{avae}-16f8 models are by far the best on reconstruction metrics, but perform on-par with FLUX.1~\cite{blackforest-flux1} when it comes to generative metrics.

To further our comparison between our best \glspl{avae} and FLUX.1~\cite{blackforest-flux1}, we compute the joint differential entropy of each model's latent space. We find that on our medical task, FLUX.1 achieves 164 nats, while \gls{avae}-4f4 achieves 1456 nats and \gls{avae}-16f8 reaches 1695 nats. The nats number measures the information entropy (or uncertainty) contained in a set. A higher nats mean that the model covers a larger data range in the latent space. These numbers show that the echocardiogram images in the FLUX.1 latent space are all encoded over a very small area of the latent space, while in the \gls{avae}-4f4 and \gls{avae}-16f8 models, the echocardiograms cover a much larger space. This is to be expected, as FLUX.1 \gls{avae} was trained on a larger diversity of images, and thus groups similar-looking images in a fraction of the space. This also mean that training a generative model on the FLUX.1 latent space constrains the model to a small subspace. This is not desirable and justifies using custom \glspl{avae} when working on domain-specific data. Therefore, for later experiments, we only consider our \gls{avae}-4f4 and \gls{avae}-16f8 models.

\subsection{Latent Image Flow Matching Models}
\subsubsection{Data preparation}
The \gls{lifm} models are trained on latent spaces. We pre-encode all the training videos with the \gls{avae}-4f4 and \gls{avae}-16f8 models, resulting in two latent datasets per real dataset.
Additionally, instead of saving the latent space $z$ (see \cref{eq:vae_reparam}), we save the outputs of the encoder $\mu$ and $\sigma$ (see \cref{eq:avae_encode}). This allows us to sample from the latent representation of each image during training, thus increasing the diversity and robustness of our generative models.

\subsubsection{Hyperparameters}
We train $15$ different \gls{lifm} models, where we vary four hyperparameters, (1) UNet~\cite{ho2020denoising} and Transformer (Flow Matching Image Transformer (FMiT))~\cite{peebles2023scalable} backbone architectures, (2) $28\times28\times4$ and $14\times14\times16$ latent space dimensions, (3) \underline{S}mall, \underline{B}ase and \underline{L}arge model sizes, (4) patch size $2$ or $4$ for the FMiT-4f4 model.

\subsubsection{Training}
The models are trained on $8\times$A40 GPUs with for a total batch size of $1\,024$. The models are trained for $1\,000\,000$ steps with a learning rate of $5e-5$, BF16 precision, and the training time varies from $640$ to $3\,600$ GPU-hours.

\subsubsection{Baselines}
We compare our image generation models with the \gls{lidm} from~\cite{reynaud2024echonet-synthetic}. This model relies on diffusion to train and sample from the model, and uses a latent space with resolution $14\times14\times4$.

\subsubsection{Evaluation}
We report the metrics for the \gls{lifm} models and the \gls{lidm}~\cite{reynaud2024echonet-synthetic} in \Cref{table:lifm}. All our models are sampled for 100 steps using the Euler ODE sampling method.

\begin{table*}[t]
\centering
\caption{Performance of the Image Generation Models.}

\label{table:lifm}
\resizebox{1.0\textwidth}{!}{
\begin{tabular}{lcrrrccccccccc}
\toprule
\multirow{2}{*}{\textbf{Model}} & 
\multirow{2}{*}{\textbf{Latent Res.}} &
\multirow{2}{*}{\textbf{Param.}} &
\multirow{2}{*}{\textbf{FLOPs}} &
\multirow{2}{*}{\textbf{Inference}} &
\multicolumn{2}{c}{\textbf{Dynamic (A4C)}} &
\multicolumn{2}{c}{\textbf{Pediatric (A4C)}} &
\multicolumn{2}{c}{\textbf{Pediatric (PSAX)}} &
\multicolumn{2}{c}{\textbf{LVH (PLAX)}} \\
\cmidrule(lr){6-7}\cmidrule(lr){8-9}\cmidrule(lr){10-11}\cmidrule(lr){12-13}
& & & & & FID$\downarrow$ & IS$\uparrow$ & FID$\downarrow$ & IS$\uparrow$ & FID$\downarrow$ & IS$\uparrow$ & FID$\downarrow$ & IS$\uparrow$ \\
\midrule

LIDM-4f8~\cite{reynaud2024echonet-synthetic} & 
$14\times14\times4$ & 
74\,M& 
-- & 
-- & 
17.3 & $2.42_{\pm0.02}$ & 
13.7 & $2.86_{\pm0.03}$ & 
16.8 & $3.05_{\pm0.03}$ & 
-- & -- \\

UNet-S-16f8 & 
$14\times14\times16$ & 
50\,M& 
1.30\,G& 
37.0\,ms& 
13.35 & $2.49_{\pm0.01}$ & 
28.52 & $2.93_{\pm0.01}$ & 
16.75 & $2.76_{\pm0.02}$ & 
22.41 & $2.26_{\pm0.02}$ \\

UNet-B-16f8 & 
$14\times14\times16$ & 
140\,M& 
4.22\,G& 
38.1\,ms& 
8.89 & $2.54_{\pm0.02}$ & 
22.54 & $3.03_{\pm0.02}$ & 
10.21 & $2.93_{\pm0.02}$ & 
16.76 & $2.37_{\pm0.01}$ \\

UNet-L-16f8 & 
$14\times14\times16$ & 
560\,M& 
14.33\,G& 
57.3\,ms& 
8.79 & $2.56_{\pm0.02}$ & 
24.05 & $3.05_{\pm0.03}$ & 
9.87 & $2.92_{\pm0.03}$ & 
16.21 & $2.34_{\pm0.01}$ \\

UNet-S-4f4 & 
$28\times28\times4$ & 
50\,M& 
5.16\,G& 
48.8\,ms& 
7.47 & $2.53_{\pm0.02}$ & 
9.51 & $3.01_{\pm0.03}$ & 
6.70 & $3.04_{\pm0.02}$ & 
14.10 & $2.43_{\pm0.02}$ \\

UNet-B-4f4 & 
$28\times28\times4$ & 
140\,M& 
14.31\,G& 
79.6\,ms& 
5.02 & $2.55_{\pm0.02}$ & 
6.63 & $3.02_{\pm0.02}$ & 
6.24 & 3.05$_{\pm0.03}$ &
11.66 & $2.48_{\pm0.02}$ \\

UNet-L-4f4 & 
$28\times28\times4$ & 
560\,M& 
57.16\,G& 
173.1\,ms& 
\textbf{4.47} & \textbf{2.58}$_{\pm0.01}$ & 
\textbf{6.22} & \textbf{3.06}$_{\pm0.01}$ &
$6.15$ & $\textbf{3.14}_{\pm0.03}$ & 
$11.19$ & $2.45_{\pm0.01}$ \\

FMiT-S/2-16f8 & 
$14\times14\times16$ & 
36\,M& 
1.06\,G& 
21.9\,ms& 
11.11 & $2.49_{\pm0.02}$ & 
33.19 & $2.87_{\pm0.02}$ & 
17.91 & $2.77_{\pm0.02}$ & 
11.99 & $2.34_{\pm0.02}$ \\

FMiT-B/2-16f8 & 
$14\times14\times16$ & 
146\,M& 
4.22\,G& 
23.4\,ms& 
10.08 & $2.50_{\pm0.02}$ & 
31.19 & $2.94_{\pm0.02}$ & 
14.35 & $2.78_{\pm0.03}$ & 
11.33 & $2.34_{\pm0.01}$ \\

FMiT-L/2-16f8 & 
$14\times14\times16$ & 
512\,M& 
15.0\,G& 
39.6\,ms& 
9.69 & $2.47_{\pm0.02}$ & 
23.62 & $3.04_{\pm0.02}$ & 
10.90 & $2.97_{\pm0.02}$ & 
11.36 & $2.33_{\pm0.02}$ \\

FMiT-S/4-4f4 & 
$28\times28\times4$ & 
36\,M& 
1.06\,G& 
23.3\,ms& 
12.43 & $2.48_{\pm0.03}$ & 
22.84 & $2.88_{\pm0.02}$ & 
17.43 & $2.82_{\pm0.01}$ & 
13.43 & $2.44_{\pm0.02}$ \\

FMiT-B/4-4f4 & 
$28\times28\times4$ & 
146\,M& 
4.22\,G& 
20.8\,ms& 
11.34 & $2.50_{\pm0.02}$ & 
15.63 & $2.94_{\pm0.02}$ & 
13.11 & $2.87_{\pm0.02}$ & 
12.15 & $2.46_{\pm0.02}$ \\

FMiT-L/4-4f4 & 
$28\times28\times4$ & 
512\,M& 
15.0\,G& 
51.0\,ms& 
8.52 & $2.54_{\pm0.02}$ & 
9.95 & $2.98_{\pm0.03}$ & 
8.12 & $3.05_{\pm0.02}$ & 
11.46 & $2.46_{\pm0.02}$ \\

FMiT-S/2-4f4 & 
$28\times28\times4$ & 
36\,M& 
4.19\,G& 
22.5\,ms& 
8.09 & $2.52_{\pm0.02}$ & 
16.02 & $2.95_{\pm0.02}$ & 
9.10 & $3.00_{\pm0.02}$ & 
7.65 & $2.51_{\pm0.02}$ \\

FMiT-B/2-4f4 & 
$28\times28\times4$ & 
146\,M& 
16.72\,G& 
46.3\,ms& 
7.50 & $2.52_{\pm0.02}$ & 
12.88 & $2.97_{\pm0.02}$ & 
7.51 & $3.02_{\pm0.03}$ & 
7.41 & $2.51_{\pm0.01}$ \\

FMiT-L/2-4f4 & 
$28\times28\times4$ & 
512\,M& 
59.42\,G& 
135.3\,ms& 
6.79 & $2.53_{\pm0.03}$ & 
8.18 & $3.01_{\pm0.02}$ & 
\textbf{5.78} & $3.07_{\pm0.02}$ & 
\textbf{6.76} & \textbf{2.51}$_{\pm0.01}$ \\

\bottomrule
\end{tabular}
}
\end{table*}

From these experiments, we observe that performance scales with the number of parameters of the models.
Our UNet-S-16f8 outperforms the LIDM-4f8~\cite{reynaud2024echonet-synthetic}, while having a similar architecture, parameter count and latent spatial resolution. This improvement comes from our improved \glspl{avae}, training hyperparameters and increased channel count.
The increased spatial resolution of 4f4 models comes at a significant computational cost (FLOPs) and inference time.
We can see that the UNet-L-4f4 and FMiT-L/2-4f4 perform similarly, with a slight edge for the UNet.

We find that a higher spatial latent resolution yields better performance, but increases computational cost, marking a clear trade-off between image quality and inference speed.

\subsection{Latent Video Flow Matching Models}
\subsubsection{Data preparation}
The \gls{lvfm} models are trained on the same latent space with the same sampling strategy as the \gls{lifm} models.
We sample a latent video and its corresponding \gls{ef} score, and extract one random frame from the video sequence. Then, we extract a 2s sample from the video, and resample it to $64$ frames. The \gls{ef} score and the random frame acts as our input conditionings, while the $64$-frame video acts as our target. 2s ensure that we cover at least one entire heartbeat.

\subsubsection{Hyperparameters}
We train 5 \gls{lvfm} models, varying the same hyperparameters as in the \gls{lifm} experiments, but limited to the \underline{S}mall variants, due to computational constraints. We explore a Spatio-Temporal UNet (STUnet)~\cite{blattmann2023stable} as well as a Spatio-Temporal Transformer (Flow Matching video Transformer (FMvT))~\cite{ma2024latte}. We also explore two latent space resolutions ($28\times28\times4$ and $14\times14\times16$), and the impact of the patch size over the FMvT-4f4 performance. Given the scaling laws observed over the \gls{lifm} model, larger \gls{lvfm} models would perform better.

\subsubsection{Training}
\gls{lvfm} models are expensive to train, as the additional temporal dimension acts as a batch size increase.
The models are trained for $1\,000\,000$ steps with a learning rate of $1e-4$ and a total batch size of $512$. The number of GPUs depends on the model size and varies from 8 to 64 H100 GPUs. The total training time ranges from $1\,500$ to $20\,000$ GPU-hours.

\subsubsection{Baselines}
We compare with two other works that explored echocardiogram synthesis over the \gls{dyn} datasets, namely EchoDiffusion~\cite{reynaud2023feature-conditioned} and EchoSyn~\cite{reynaud2024echonet-synthetic} (\gls{lvdm}-4f8). We note that EchoDiffusion is a pixel-space diffusion model, and \gls{lvdm}-4f8 is a latent diffusion model.

\subsubsection{Evaluation}

We present the metrics in \Cref{table:lvfm}. The models are sampled for 100 steps with an Euler ODE sampler. 

\begin{table*}[t]
\centering
\caption{Performance Comparison of the Video Generation Models}
\label{table:lvfm}
\resizebox{1.0\textwidth}{!}{
\begin{tabular}{lcrrrcccccccccccc}
\toprule
\multirow{2}{*}{\textbf{Model}} & \multirow{2}{*}{\textbf{Latent Res.}} & \multirow{2}{*}{\textbf{Params.}} & \multirow{2}{*}{\textbf{FLOPs}} & \multirow{2}{*}{\textbf{Inference}} & \multicolumn{3}{c}{\textbf{Dynamic (A4C)}} & \multicolumn{3}{c}{\textbf{Pediatric (A4C)}} & \multicolumn{3}{c}{\textbf{Pediatric (PSAX)}} & \multicolumn{3}{c}{\textbf{LVH (PLAX)}} \\
\cmidrule(lr){6-8}\cmidrule(lr){9-11}\cmidrule(lr){12-14}\cmidrule(lr){15-17}
& & & & & FID$\downarrow$ & FVD$_{16}$$\downarrow$ & IS$\uparrow$ & FID$\downarrow$ & FVD$_{16}$$\downarrow$ & IS$\uparrow$ & FID$\downarrow$ & FVD$_{16}$$\downarrow$ & IS$\uparrow$ & FID$\downarrow$ & FVD$_{16}$$\downarrow$ & IS$\uparrow$ \\
\midrule
EchoDiffusion~\cite{reynaud2023feature-conditioned}& $64\times112\times112$& 95\,M & --        & 279.00\,s& $24.0$  &   $228$ & $\textbf{2.59}_{\pm0.06}$ & --      & --      & --               & --      & --      & --               & -- & -- & --\\
LVDM-4f8~\cite{reynaud2024echonet-synthetic} & $64\times14\times14\times4$   & 144\,M    & --        & 2.40\,s& $17.4$  &  $71.4$ & $2.31_{\pm0.08}$ & $24.8$  & $112.2$ & $2.69_{\pm0.18}$ & $33.0$  & $126.9$ & $2.49_{\pm0.09}$ & -- & -- & -- \\
STUNet-S-16f8                      & $64\times14\times14\times16$  & 46\,M     &  79.86\,G & 1.92\,s & $23.81$ & $173.0$ & $2.21_{\pm0.05}$ & $36.47$ & $317.0$ & $2.50_{\pm0.07}$ & $57.36$ & $379.4$ & $2.49_{\pm0.08}$ & $30.80$ & $215.6$ & $2.23_{\pm0.06}$ \\
STUNet-S-4f4                       & $64\times28\times28\times4$   & 46\,M     & 317.19\,G & 4.80\,s & $13.73$ &  $40.0$ & $2.40_{\pm0.06}$ & $19.44$ & $107.6$ & $2.78_{\pm0.12}$ & $26.38$ & $135.3$ & $2.87_{\pm0.09}$ & $14.14$ & $49.4$ & $2.53_{\pm0.10}$ \\
FMvT-S/2-16f8                      & $64\times14\times14\times16$  & 36\,M     & 103.51\,G & 0.71\,s & $13.83$ &  $85.9$ & $2.36_{\pm0.04}$ & $21.69$ & $195.5$ & $2.71_{\pm0.11}$ & $34.62$ & $218.2$ & $2.72_{\pm0.11}$ & $15.33$ & $74.5$ & $2.43_{\pm0.07}$ \\
FMvT-S/4-4f4                       & $64\times28\times28\times4$   & 36\,M     & 103.51\,G & 0.74\,s & $14.78$ &  $48.8$ & $2.38_{\pm0.05}$ & $21.05$ & $153.8$ & $2.80_{\pm0.13}$ & $31.64$ & $181.2$ & $2.78_{\pm0.08}$ & $14.80$ & $46.9$ & $2.50_{\pm0.09}$ \\
FMvT-S/2-4f4                       & $64\times28\times28\times4$   & 36\,M     & 430.31\,G & 2.94s & $\textbf{7.92}$ & $\textbf{28.1}$ & ${2.46}_{\pm0.06}$ & $\textbf{12.26}$ & $\textbf{75.4}$ & $\textbf{2.88}_{\pm0.14}$ & $\textbf{16.77}$ & $\textbf{84.6}$ & $\textbf{2.99}_{\pm0.09}$ & $\textbf{7.71}$ & $\textbf{23.2}$ & $\textbf{2.58}_{\pm0.06}$ \\
\bottomrule
\end{tabular}
}
\end{table*}

\begin{figure}[!h]
        \centering
    \includegraphics[width=1.0\linewidth]{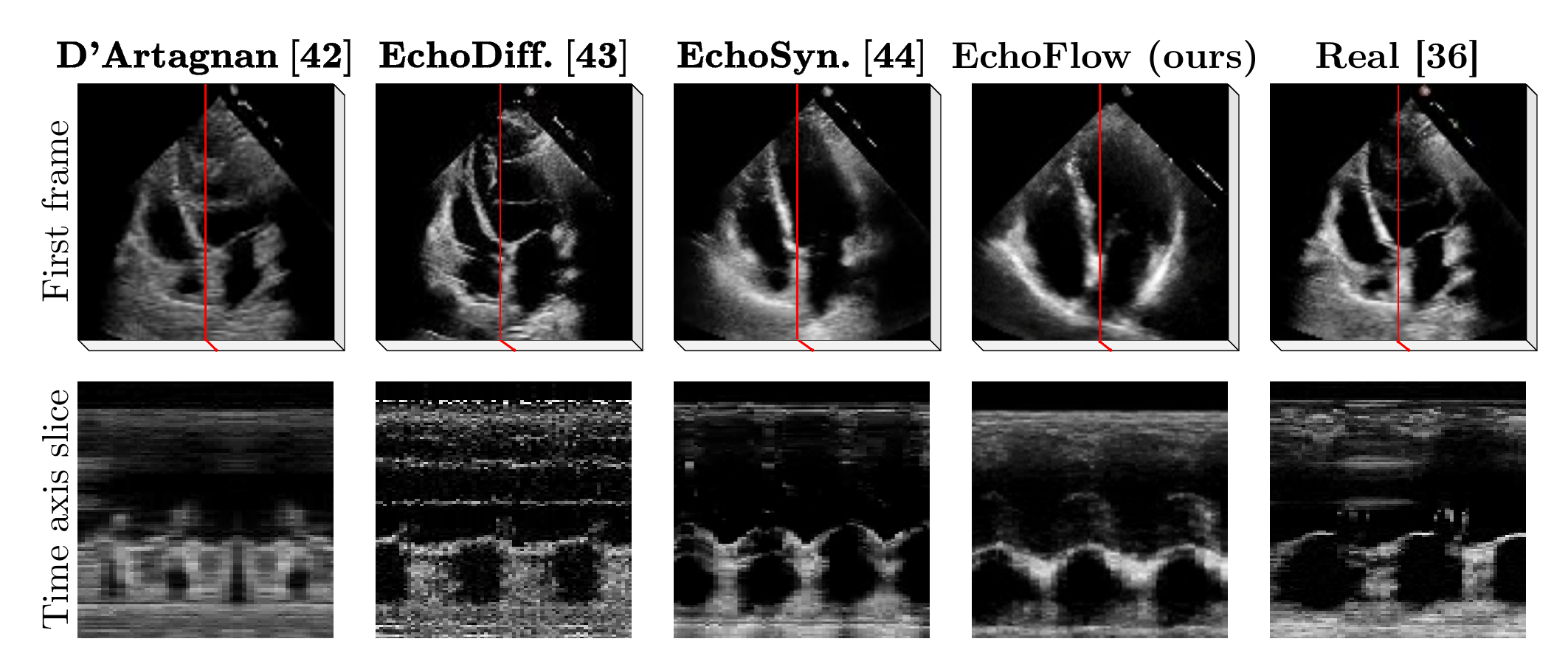}
    \caption{Qualitative comparison of recent echocardiogram synthesis methods in spatial and temporal domain.}
    \label{fig:qualitative}
\end{figure}

Based on the results in \Cref{table:lvfm}, we see that FMvT outperforms the STUNet~\cite{blattmann2023stable}. We believe that this is due to the higher performance of the temporal attention modules and increased stability of FMvT, observed during training.
Despite all models having a similar parameter count, FMvT-S/2-4f4 beats all other model by a significant margin. It is also the most computationally expensive model, because it operates on a higher resolution latent space. 
%
By picking a better latent space dimension and training the model longer, FMvT-S/2-4f4 beats all previous state-of-the-art models while being $2-3\times$ smaller.
%
When sampling the FMvT-S/2-4f4 model we reach optimal performance when using 100 steps, according to \Cref{table:sampling_steps}, while keeping the sampling time similar to the previous state-of-the-art model~\cite{reynaud2024echonet-synthetic}.
%
Based on these results, we select FMvT-S/2-4f4 as our main video generation model for all further experiments.
We visually compare models in \Cref{fig:qualitative}.

\subsubsection{Sampling Parameters}

We evaluate the impact of sampling steps over the FMvT-S/2-4f4 performance in \Cref{table:sampling_steps}. We observe that the results with 25 sampling steps outperform the previous state-of-the-art~\cite{reynaud2024echonet-synthetic}, but we hit the point of diminishing returns at 100 sampling steps. Therefore, we use a 100 sampling steps.
\begin{table*}[t]
\centering
\begin{minipage}[t]{0.68\textwidth}
    \centering
    \caption{Video generation performance per sampling steps.}
    \label{table:sampling_steps}
    \resizebox{\textwidth}{!}{%
    \begin{tabular}{cccccccccccccc}
        \toprule
        \multirow{2}{*}{\textbf{Steps}} & \multirow{2}{*}{\textbf{Inference}} & \multicolumn{3}{c}{\textbf{Dynamic (A4C)}} & \multicolumn{3}{c}{\textbf{Pediatric (A4C)}} & \multicolumn{3}{c}{\textbf{Pediatric (PSAX)}} & \multicolumn{3}{c}{\textbf{LVH (PLAX)}} \\
        \cmidrule(lr){3-5}\cmidrule(lr){6-8}\cmidrule(lr){9-11}\cmidrule(lr){12-14}
        & & FID$\downarrow$ & FVD$_{16}$$\downarrow$ & IS$\uparrow$ & FID$\downarrow$ & FVD$_{16}$$\downarrow$ & IS$\uparrow$ & FID$\downarrow$ & FVD$_{16}$$\downarrow$ & IS$\uparrow$ & FID$\downarrow$ & FVD$_{16}$$\downarrow$ & IS$\uparrow$ \\
        \midrule
        5 & 0.26s & $56.62$ & $374.3$ & $2.13_{\pm0.05}$ & $39.01$ & $394.0$ & $2.61_{\pm0.12}$ & $46.50$ & $496.3$ & $2.83_{\pm0.11}$ & $61.02$ & $523.5$ & $2.26_{\pm0.09}$ \\
        10 & 0.60s & $24.24$ & $126.6$ & $2.32_{\pm0.05}$ & $19.64$ & $158.5$ & $2.77_{\pm0.15}$ & $23.76$ & $164.2$ & $3.02_{\pm0.09}$ & $26.77$ & $172.5$ & $2.39_{\pm0.09}$ \\
        25 & 1.00s & $11.06$ & $42.2$ & $2.40_{\pm0.07}$ & $13.18$ & $77.8$ & $2.85_{\pm0.14}$ & $16.73$ & $82.6$ & $3.02_{\pm0.10}$ & $11.72$ & $46.9$ & $2.52_{\pm0.06}$ \\
        50 & 1.64s & $8.73$ & $30.7$ & $2.43_{\pm0.05}$ & $12.09$ & $70.6$ & $2.91_{\pm0.16}$ & $15.98$ & $81.3$ & $3.03_{\pm0.06}$ & $8.94$ & $26.1$ & $2.56_{\pm0.08}$ \\
        100 & 2.94s & $7.92$ & $28.1$ & $2.46_{\pm0.06}$ & $12.26$ & $75.4$ & $2.88_{\pm0.14}$ & $16.77$ & $84.6$ & $2.99_{\pm0.09}$ & $7.71$ & $23.2$ & $2.58_{\pm0.06}$ \\
        200 & 5.61s & $7.27$ & $28.0$ & $2.48_{\pm0.05}$ & $12.28$ & $78.3$ & $2.91_{\pm0.16}$ & $16.90$ & $90.6$ & $3.03_{\pm0.10}$ & $7.63$ & $23.4$ & $2.58_{\pm0.08}$ \\
        500 & 13.71s & $7.44$ & $29.1$ & $2.47_{\pm0.07}$ & $11.90$ & $79.2$ & $2.87_{\pm0.12}$ & $16.59$ & $92.5$ & $3.03_{\pm0.09}$ & $7.31$ & $24.6$ & $2.59_{\pm0.06}$ \\
        \bottomrule
    \end{tabular}}
\end{minipage}%
\hfill
\begin{minipage}[t]{0.3\textwidth}
    \centering
    \caption{Re-Identification model metrics.}
    \label{table:reid}
    \resizebox{\textwidth}{!}{%
    \begin{tabular}{llcc}
        \toprule
        \textbf{Dataset} & \textbf{\gls{lifm} Model} & Threshold $\tau$ & Rejection rate \\
        \midrule
        \gls{dyn} & UNet-L-4f4 & 0.9997 & $35.8\%$ \\
        \gls{ped}\,(\gls{a4c}) & UNet-L-4f4 & 0.9851 & $33.0\%$ \\
        \gls{ped}\,(\gls{psax}) & FMiT-L/2-4f4 & 0.9953 & $33.2\%$ \\
        \gls{lvh} & FMiT-L/2-4f4 & 0.9950 & $35.6\%$ \\
        \bottomrule
    \end{tabular}}
\end{minipage}%
\end{table*}

We also explore the impact of \gls{cfg} over the FMvT-S/2-4f4 model performance in \Cref{table:cfg}. 
The first row, where $\lambda_{\text{cfg}}=0$, shows the performance of the unconditioned or partially conditioned model. Indeed, $v_{unc}$ in \cref{eq:cfg_def} can be fully unconditional, or partially conditioned if we pass either the \gls{ef} or the anatomical frame to the model. By selectively passing none or one of these conditions, we orient the \gls{cfg} process in different directions. Therefore, for experiments with $\lambda_{\text{cfg}}=0$, we see that the model performance heavily depends on the input anatomy, as removing the anatomy (rows 0 and 2), significantly degrades performance. When removing only the \gls{ef} input (row 1), the visual performance does not degrade.

The $\lambda_{\text{cfg}}=1$ row shows the baseline model, where \gls{cfg} is not applied. There is no consensus on which value to set for $\lambda_{\text{cfg}}$. Nonetheless, $\lambda_{\text{cfg}}=1$ maximises none of the metrics. The other consideration is that a higher $\lambda_{\text{cfg}}$ decreases sample diversity~\cite{ho2022classifier-free}. Therefore, we select $\lambda_{\text{cfg}}=2$ with no negative conditionings ($\varnothing$). This configuration demonstrates increased overall performance on most metrics.

\begin{table}[t]
\centering
\caption{Video generation performance per CFG weight.}
\label{table:cfg}
\resizebox{1.0\columnwidth}{!}{
\begin{tabular}{cccccccccccccc}
\toprule
\multirow{2}{*}{\shortstack{\textbf{CFG Scale} \\ \textbf{($\lambda_{\text{cfg}}$)}}} & \multirow{2}{*}{\shortstack{\textbf{Negative} \\ \textbf{Condition}}} & \multicolumn{3}{c}{\textbf{Dynamic (A4C)}} & \multicolumn{3}{c}{\textbf{Pediatric (A4C)}} & \multicolumn{3}{c}{\textbf{Pediatric (PSAX)}} & \multicolumn{3}{c}{\textbf{LVH (PLAX)}} \\
\cmidrule(lr){3-5}\cmidrule(lr){6-8}\cmidrule(lr){9-11}\cmidrule(lr){12-14}
& & FID$\downarrow$ & FVD$_{16}$$\downarrow$ & IS$\uparrow$ & FID$\downarrow$ & FVD$_{16}$$\downarrow$ & IS$\uparrow$ & FID$\downarrow$ & FVD$_{16}$$\downarrow$ & IS$\uparrow$ & FID$\downarrow$ & FVD$_{16}$$\downarrow$ & IS$\uparrow$ \\
\midrule
0.0  & $\varnothing$ & 44.99 & 271.21 & $\textbf{2.75}_{\pm0.06}$ & 77.71 & 554.20 & $2.75_{\pm0.06}$ & 110.59 & 785.69 & $2.75_{\pm0.06}$ & 43.00 & 326.71 & $\textbf{2.75}_{\pm0.06}$ \\          
0.0  & Anatomy       & 7.94  & 28.94  & $2.44_{\pm0.05}$ & 12.05 & 72.40  & $2.89_{\pm0.14}$ & 16.81  & 88.65  & $2.99_{\pm0.08}$ & 8.01  & 22.73  & $2.59_{\pm0.05}$ \\                    
0.0  & LVEF          & 45.47 & 277.85 & $\textbf{2.75}_{\pm0.07}$ & 76.70 & 553.17 & $2.74_{\pm0.03}$ & 108.90 & 770.57 & $2.75_{\pm0.05}$ & 40.39 & 301.29 & $2.73_{\pm0.06}$ \\
\midrule
1.0  & N/A               & 7.92 & 28.1  & $2.46_{\pm0.06}$ & 12.26 & 75.4  & $2.88_{\pm0.14}$ & 16.77 & 84.6  & $2.99_{\pm0.09}$ & 7.71  & 23.2  & $2.58_{\pm0.06}$ \\
\midrule
2.0  & $\varnothing$ & 5.62 & 51.07 & $2.49_{\pm0.06}$ & 8.61  & 65.41 & $2.91_{\pm0.19}$ & 10.43 & \textbf{62.87} & $3.15_{\pm0.11}$ & \textbf{5.44} & 58.79 & $2.58_{\pm0.11}$ \\
2.0  & Anatomy       & 8.07  & 28.57 & $2.47_{\pm0.06}$ & 11.93 & 71.65 & $2.92_{\pm0.15}$ & 16.14 & 85.21 & $3.05_{\pm0.11}$ & 7.59  & 22.27 & $2.57_{\pm0.07}$ \\                    
2.0  & LVEF          & \textbf{5.57} & 51.97 & $2.51_{\pm0.07}$ & 8.60  & 66.26 & $2.94_{\pm0.18}$ & 10.79 & 64.08 & $3.09_{\pm0.12}$ & 5.55  & 60.80 & $2.61_{\pm0.09}$ \\
\midrule
3.0  & $\varnothing$ & 5.63 & 133.47 & $2.49_{\pm0.05}$ & 8.64  & 121.44 & $2.94_{\pm0.14}$ & \textbf{10.35} & 132.03 & $3.13_{\pm0.14}$ & 6.36  & 211.90 & $2.69_{\pm0.12}$ \\
3.0  & Anatomy       & 7.65 & 27.35  & $2.47_{\pm0.05}$ & 11.93 & 68.37  & $2.85_{\pm0.12}$ & 16.20 & 82.57  & $3.06_{\pm0.10}$ & 7.65  & 21.95  & $2.59_{\pm0.09}$ \\
3.0  & LVEF          & 5.90 & 138.80 & $2.51_{\pm0.08}$ & 8.56  & 126.21 & $\textbf{2.95}_{\pm0.16}$ & 10.42 & 131.16 & $\textbf{3.20}_{\pm0.13}$ & 6.47  & 217.02 & $2.63_{\pm0.10}$ \\
\midrule
4.0  & $\varnothing$ & 6.00 & 217.35 & $2.48_{\pm0.08}$ & \textbf{8.48} & 198.08 & $2.93_{\pm0.14}$ & 10.68 & 192.71 & $3.17_{\pm0.09}$ & 6.65  & 365.51 & $2.66_{\pm0.09}$ \\
4.0  & Anatomy       & 7.59 & 27.22  & $2.50_{\pm0.06}$ & 13.04 & 68.25  & $2.91_{\pm0.15}$ & 15.74 & 80.70  & $3.03_{\pm0.10}$ & 7.42  & 22.34  & $2.58_{\pm0.06}$ \\
4.0  & LVEF          & 6.07 & 222.84 & $2.53_{\pm0.11}$ & 8.64  & 199.98 & $\textbf{2.95}_{\pm0.19}$ & 10.44 & 202.98 & $3.11_{\pm0.11}$ & 6.92  & 345.66 & $2.60_{\pm0.08}$ \\
\midrule
5.0  & $\varnothing$ & 6.14 & 250.22 & $2.56_{\pm0.09}$ & 8.91 & 256.64 & $2.94_{\pm0.17}$ & 10.76 & 272.65 & $3.11_{\pm0.09}$ & 7.09  & 445.77 & $2.68_{\pm0.09}$ \\
5.0  & Anatomy       & 7.43 & 26.21  & $2.48_{\pm0.06}$ & 11.95 & 64.77  & $2.90_{\pm0.14}$ & 15.23 & 80.92  & $3.04_{\pm0.10}$ & 7.28  & 22.22  & $2.58_{\pm0.06}$ \\
5.0  & LVEF          & 6.26 & 262.29 & $2.52_{\pm0.09}$ & 9.20  & 257.50 & $2.92_{\pm0.19}$ & 10.61 & 257.38 & $3.15_{\pm0.14}$ & 6.96  & 450.39 & $2.65_{\pm0.08}$ \\
\midrule
10.0 & $\varnothing$ & 7.14 & 200.20 & $2.59_{\pm0.06}$ & 9.78  & 266.35 & $2.92_{\pm0.16}$ & 11.58 & 297.65 & $3.09_{\pm0.09}$ & 7.34  & 416.93 & $2.72_{\pm0.07}$ \\
10.0 & Anatomy       & 7.31 & \textbf{25.36} & $2.51_{\pm0.04}$ & 11.66 & \textbf{59.93} & $2.91_{\pm0.14}$ & 14.47 & 71.00  & $3.12_{\pm0.13}$ & 6.75  & \textbf{21.43} & $2.59_{\pm0.08}$ \\
10.0 & LVEF          & 7.26 & 213.78 & $2.58_{\pm0.09}$ & 9.72  & 262.52 & $2.88_{\pm0.12}$ & 11.61 & 299.47 & $3.18_{\pm0.11}$ & 7.33  & 413.39 & $2.69_{\pm0.06}$ \\
\bottomrule
\end{tabular}
}
\end{table}


\subsection{Re-Identification Models}
\subsubsection{Data preparation}
To train the \gls{reid} models, we use the real latent datasets, encoded with \gls{avae}-4f4. We follow the method described in \cref{subsec:reid} for sampling.

\subsubsection{Hyperparameters}
We use a ResNet18 model, which takes as input a single latent image with resolution $28\times28\times4$ and projects it to a 1D vector of length $256$. All four \gls{reid} models are configured with these parameters.

\subsubsection{Training}
We train one {\gls{reid} model per dataset. The models are trained over latent space images and thus have very low compute requirements. Each model is trained for $50\,000$ steps, as the accuracy reaches $99\%$ on the validation set. The learning rate is set to $1e-4$ and the batch size to $32$.

\subsubsection{Evaluation}
After training the \gls{reid} models, we compute $\tau$ and apply each model on $100\,000$ latent images generated with the corresponding \gls{lifm}. For all datasets, the memorization rate is around $34\%$, as shown in \Cref{table:reid}.

\subsection{EchoFlow Dataset}

For each dataset, we have identified the best \gls{avae}, \gls{lifm} and \gls{lvfm} models. We have also trained \gls{reid} models on the corresponding real latent datasets.
Putting everything together, the \gls{lifm} generates $100\,000$ samples per dataset and we filter them with their corresponding \gls{reid} model. This gives us two sets of images, an unfiltered and a filtered one. To enable fair comparison with the real datasets, we randomly select 
synthetic images in the same quantities as the real datasets and respect the original splits sizes, as listed in \Cref{table:echoflow}. 
As such, we select the right amount of images from the unfiltered synthetic image collections for each dataset and split.
Then, we duplicate these sets, remove any non-privacy compliant image, and replace them with unused privacy compliant images, forming our privacy-compliant sets.
All images are then animated with the \gls{lvfm}. We consider the average number of frames in each real dataset to decide how many times each anatomy should be animated to obtain a similar amount of frames per anatomy. As our video model generates 64-frame videos, we generate 3 videos per synthetic image for \gls{dyn} and \gls{lvh}, and 2 for \glspl{ped}. \gls{ef} scores are randomly sampled in the range $10\%-90\%$ and paired with a synthetic image. 

\begin{table}[t]
\centering
\caption{Datasets size and selected model. A.F.C stands for `Average Frame Count'.}
\label{table:echoflow}
\resizebox{1.0\columnwidth}{!}{
\begin{tabular}{lllcccc}
\toprule
\textbf{Dataset}         & 
\textbf{\gls{lifm} Model}& 
\textbf{\gls{lvfm} Model}& 
\textbf{Train}           & 
\textbf{Val.}            & 
\textbf{Test}            &
\textbf{A.F.C.} \\
\midrule
\gls{dyn}               & UNet-L-4f4    & FMvT-S/2-4f4 & 7465 & 1288 & 1277 &  177\\
\gls{ped}\,(\gls{a4c})  & UNet-L-4f4    & FMvT-S/2-4f4 & 2580 & 336  & 368  &  101\\
\gls{ped}\,(\gls{psax}) & FMiT-L/2-4f4  & FMvT-S/2-4f4 & 3559 & 448  & 519  &  103\\
\gls{lvh}               & FMiT-L/2-4f4  & FMvT-S/2-4f4 & 9508 & 1076 & 332  &  165\\
\bottomrule
\end{tabular}
}
\end{table}

In order to ensure the best possible results, we relabel every video with a pre-trained \gls{ef} regression model. Each video gets attributed this new \gls{ef} score as its label. This is done to correct approximations introduced by the \gls{lvfm} model.

By following these steps, we generate a total of 8 video collections: a non-privacy-compliant (NPC) and a privacy-compliant (PC) synthetic version of each original dataset.

\subsection{Evaluation on Downstream Tasks}

We compare the performance of an \gls{ef} regression model trained on various datasets in \Cref{table:downstream}. We report the results claimed in each dataset's original work (`Claimed') and the results we obtained from reproducing said work (`Reproduced') row.
We then explore the impact of training the regression model on datasets reconstructed with our \glspl{avae} and observe no performance degradation compared to the Reproduced results.
For the synthetic datasets, we compare our datasets (EchoFlow) with existing baselines (EchoDiff.~\cite{reynaud2023feature-conditioned} and EchoSyn.~\cite{reynaud2024echonet-synthetic}). In a controlled and fair setting, our new synthetic datasets outperform all previous approaches and reach performance parity with the real datasets, even on the privacy-preserving synthetic datasets.

\begin{table}[t]
\centering
\caption{Downstream tasks performance.}
\label{table:downstream}
\resizebox{1.0\columnwidth}{!}{
\begin{tabular}{lcccccccccccc}
\toprule
\multirow{2}{*}{\textbf{}} & \multicolumn{3}{c}{\textbf{Dynamic (A4C)}} & \multicolumn{3}{c}{\textbf{Pediatric (A4C)}} & \multicolumn{3}{c}{\textbf{Pediatric (PSAX)}} & \multicolumn{3}{c}{\textbf{LVH (PLAX)}} \\
\cmidrule(lr){2-4}\cmidrule(lr){5-7}\cmidrule(lr){8-10}\cmidrule(lr){11-13}
& {R2}$\ua$ & {MAE}$\da$ & {RMSE}$\da$ & {R2}$\ua$ & {MAE}$\da$ & {RMSE}$\da$ & {R2}$\ua$ & {MAE}$\da$ & {RMSE}$\da$ & {R2}$\ua$ & {MAE}$\da$ & {RMSE}$\da$ \\
\midrule
Claimed         & 0.81 & 4.05 & 5.32 & 0.70 & 4.15 & 5.70 & 0.74 & 3.80 & 5.14 &  --  &  --  &  --  \\
Reproduced      & 0.81 & 3.98 & 5.29 & 0.68 & 4.19 & 5.71 & 0.71 & 3.79 & 5.14 & 0.55 & 6.42 & 8.38 \\
\gls{avae}-4f4 Rec.    & 0.81 & 4.08 & 5.38 & 0.70 & 4.65 & 6.37 & 0.68 & 4.21 & 5.95 & 0.57 & 6.43 & 8.21 \\
\gls{avae}-16f8 Rec.   & 0.81 & 3.99 & 5.28 & 0.68 & 4.57 & 6.52 & 0.70 & 4.12 & 5.78 & 0.58 & 6.36 & 8.13 \\
\midrule
EchoDiff.~\cite{reynaud2023feature-conditioned} & 0.55 & 6.02 & 8.21 &  --  &  --  &  --  &  --  &  --  &  --  &  --  &  --  &  -- \\
EchoSyn.~\cite{reynaud2024echonet-synthetic}  & 0.75 & 4.55 & 6.10 & 0.70 & 5.06 & 7.07 & 0.68 & 4.82 & 6.95 &  --  &  --  &  --  \\
EchoFlow (NPC.)                    & 0.81 & 4.10 & 5.40 & 0.72 & 4.08 & 6.10 & 0.73 & 3.92 & 5.48 & 0.53 & 6.62 & 8.60 \\
EchoFlow (PC.)                     & 0.81 & 4.05 & 5.34 & 0.72 & 4.29 & 6.09 & 0.72 & 3.95 & 5.60 & 0.55 & 6.54 & 8.34 \\
\bottomrule
\end{tabular}
}
\end{table}

\section{Conclusion}

In this work, we introduced a comprehensive framework for generating high-fidelity, privacy-preserving cardiac ultrasound images and videos. By leveraging generative modelling techniques, such as Adversarial Variational Auto-Encoders for domain-specific latent space construction, Latent Flow Matching for both image and video generation, and a robust Re-Identification mechanism for privacy filtering, we demonstrated that privacy preserving,  synthetic medical imaging datasets can achieve performance parity with real data in downstream clinical tasks like \gls{ef} regression. Our extensive experiments across multiple datasets (including EchoNet-Dynamic, EchoNet-Pediatric, and EchoNet-LVH) reveal that scaling model size and training time effectively closes the quality and diversity gap between real and synthetic samples. This work not only paves the way for broader data sharing in medical imaging without compromising patient privacy but also establishes a solid foundation for future research in generative models across diverse clinical applications.

\bibliographystyle{splncs04}
\bibliography{bibshort}
\end{document}